%% file: main.tex
\pdfoutput=1

\documentclass[11pt]{article}

\usepackage{acl}

\usepackage{times}
\usepackage{latexsym}

\usepackage[T1]{fontenc}

\usepackage[utf8]{inputenc}

\usepackage{microtype}

\usepackage{booktabs}
\usepackage{amsmath}
\usepackage{cleveref}
\crefformat{section}{\S#2#1#3}
\crefformat{subsection}{\S#2#1#3}
\crefformat{subsubsection}{\S#2#1#3}
\usepackage{graphicx}
\usepackage{CJKutf8}
\usepackage{multirow}

\usepackage{algorithm,algpseudocode}

\newcommand{\minisection}[1]{\noindent{\bf {#1}.}}
\newcommand{\minisectionNoDot}[1]{\noindent{\bf {#1}}}

\newcommand{\Appendix}[1]{Appendix \ref{#1}}
\newcommand{\Table}[1]{Table \ref{#1}}
\newcommand{\Figure}[1]{Figure \ref{#1}}
\newcommand{\Algorithm}[1]{Algorithm \ref{#1}}
\newcommand{\eg}{{\it e.g.}}
\newcommand{\ie}{{\it i.e.}}
\newcommand{\sentence}[1]{``{\it #1}''}
\newcommand{\word}[1]{{\it #1}}
\newcommand{\token}[1]{``\texttt{#1}''}

\newcommand{\package}[1]{\texttt{#1}}

\usepackage{mwe}
\usepackage{subfig}

\newcommand{\papertitle}{Pretraining with Artificial Language: Studying Transferable Knowledge in Language Models}
\title{Pretraining with Artificial Language: \\Studying Transferable Knowledge in Language Models}

\author{Ryokan Ri and Yoshimasa Tsuruoka \\
  The University of Tokyo\\
  7-3-1 Hongo, Bunkyo-ku, Tokyo, Japan \\
  {\tt \{li0123,tsuruoka\}@logos.t.u-tokyo.ac.jp} \\}

\begin{document}
\maketitle

\input{sections/macros}

\begin{abstract}
\input{sections/00-abstract}
\end{abstract}

\input{sections/01-introduction}
\input{sections/05-related_work}
\input{sections/02-approach}
\input{sections/03-experiment-tilt}
\input{sections/04-experiment-bert}
\input{sections/04.5-probing}
\input{sections/06-discussion}

\newpage
\section*{Acknowledgement}
We thank the anonymous reviewers for their insightful comments and constructive suggestions to improve the paper.

 \bibliography{references/references}

 \bibliographystyle{acl_natbib}

\clearpage
\input{sections/99-appendix}
\end{document}

%% file: sections/macros.tex
\newcommand{\Lone}{L1}
\newcommand{\Ltwo}{L2}

\newcommand{\FromScratch}{From scratch}
\newcommand{\RandomWeights}{Random weights}
\newcommand{\Uniform}{Uniform}
\newcommand{\Zipf}{Zipf}
\newcommand{\LogLinear}{Log-linear}
\newcommand{\FlatParenthesis}{Flat Parenthesis}
\newcommand{\NestingParenthesis}{Nesting Parenthesis}
\newcommand{\FlatDependency}{Flat Dependency}
\newcommand{\NestingDependency}{Nesting Dependency}

\newcommand{\Dependency}{Dependency}
\newcommand{\Parenthesis}{Parenthesis}
\newcommand{\Flat}{Flat}
\newcommand{\Nesting}{Nesting}
\newcommand{\LSTM}{LSTM}
\newcommand{\Transformer}{Transformer}

\newcommand{\Spanish}{Spanish}
\newcommand{\Japanese}{Japanese}
\newcommand{\English}{English}

\newcommand{\PoSTagging}{PoS tagging}
\newcommand{\DependencyParsing}{dependency parsing}

%% file: sections/00-abstract.tex
We investigate what kind of structural knowledge learned in neural network encoders is transferable to processing natural language.
We design \emph{artificial languages} with structural properties that mimic natural language, pretrain encoders on the data, and see how much performance the encoder exhibits on downstream tasks in natural language.
Our experimental results show that pretraining with an artificial language with a nesting dependency structure provides some knowledge transferable to natural language.
A follow-up probing analysis indicates that its success in the transfer is related to the amount of encoded contextual information and what is transferred is the knowledge of \emph{position-aware context dependence} of language.
Our results provide insights into how neural network encoders process human languages and the source of cross-lingual transferability of recent multilingual language models.

%% file: sections/01-introduction.tex
\section{Introduction}

\begin{figure}[t]
  \begin{center}
    \includegraphics[width=7cm]{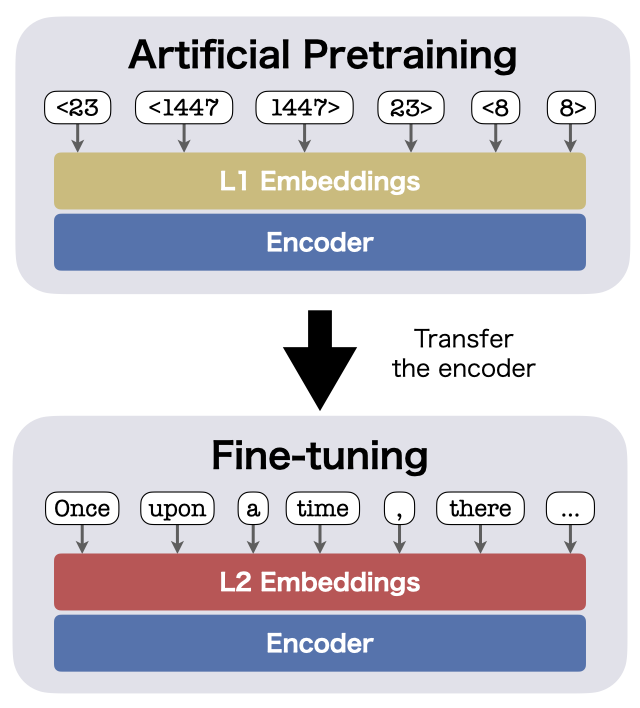}
    \caption{Transfer from artificial language to natural language. The artificial language encodes some structural properties (\eg{}, token distributions, dependency structures) and we study how the learning of such properties can be transferred to natural language.}
    \label{fig:artificial_pretraining}
  \end{center}
\end{figure}

Pretrained language models \citep{devlin2018bert,Yang2019XLNet:Understanding,JMLR:v21:20-074} have demonstrated strong empirical performance not only within a language but also \emph{across} languages.
Language models pretrained with a mix of monolingual corpora, such as multilingual BERT, exhibit a decent zero-shot cross-lingual transfer capability, \ie{}, a model fine-tuned in a single source language (\Lone{}) can solve the task in another language (\Ltwo{}) \citep{conneau-etal-2020-unsupervised,xue-etal-2021-mt5}.
Surprisingly, the transfer happens without lexical overlaps between \Lone{} and \Ltwo{} \citep{K-mBERT-ICLR-2020,conneau-etal-2020-emerging} or even without joint pretraining \citep{artetxe-etal-2020-cross}:
an encoder only pretrained on \Lone{} can be transferred to \Ltwo{} without any parameter updates.
These results suggest that, whether the encoder is trained on single or multiple languages, it learns some {\it transferable knowledge} about language.

However, the characteristics of such transferable knowledge are still underexplored.
Recent studies with the probing methodology \citep{ijcai2018-796,conneau-etal-2018-cram} have revealed that multilingual BERT captures language-independent linguistic structures such as universal dependency relations \citep{chi-etal-2020-finding} and subjecthood \citep{papadimitriou-etal-2021-deep}, but it remains unknown whether learning such linguistic properties actually contributes to the performance, and whether there exists more abstract knowledge transferred across languages.

In this study, we try to shed light on these questions with the framework of the Test for Inductive Bias via Language Model Transfer \citep{papadimitriou-jurafsky-2020-learning}, focusing on designing \emph{artificial languages} with natural-language-like structural properties (\Figure{fig:artificial_pretraining}).
We pretrain encoders with artificial languages and transfer the encoders to natural language tasks with their parameters frozen.
This enables us to see how learning the specific structural properties of the artificial language affects the downstream performance.

Specifically, we explore whether it is beneficial for the encoder to know the following two characteristics of natural language: word distributions and latent dependency structures.
We design artificial languages that represent such characteristics and perform an extensive study with different encoder architectures (\LSTM{} and \Transformer{}) pretraining objectives (causal and masked language modelings).

The contribution is summarized as follows:

\begin{itemize}
  \item We first start by complementing the study in \citet{papadimitriou-jurafsky-2020-learning}.
  We train \LSTM{} and \Transformer{} encoders with the sentence-level causal language modeling task and evaluate the encoders in English.
  We show that an artificial language that models simple statistical dependency within a sentence provides decent transferable knowledge on natural language modeling.
  Furthermore, we find that the inductive bias of a nesting head-to-tail dependency structure is more useful than a flat one.

  \item We then proceed to investigate transfer learning in masked language modeling \citep{devlin2018bert}, one of the current dominant pretraining paradigms.
  We evaluate pretrained \Transformer{} encoders with dependency parsing and confirm that the nesting dependency structure is important to learn the structure of natural language.

  \item We hypothesize that the transfer performance of pretrained encoders is related to the way the encoder preserves the input contextual information in the output vectors.
  We perform a probing experiment and find that the artificial language with the nesting dependency structure trains encoders to encode the information on adjacent tokens into the output vector of each token.
  We conclude this paper with the hypothesis that a part of transferable knowledge in language models could be explained by the knowledge of {\it position-aware context dependence} of language.
\end{itemize}

%% file: sections/05-related_work.tex
\section{Related Work}

\subsection{Transferable Structural Knowledge in Pretrained Encoders}
Multilingual language models trained with masked language modeling objective \citep{devlin2018bert,Doddapaneni2021APO} have demonstrated a surprisingly strong cross-lingual transfer capability \citep{liu-etal-2020-multilingual}, given the model is only trained with a mix of monolingual corpora.
This leads to several studies investigating the source of the cross-lingual capability of multilingual models.

An early common hypothesis was that the models take advantage of a common word-piece vocabulary across languages \citep{wu-dredze-2019-beto,pires-etal-2019-multilingual}, which provides cross-lingual alignment signals to learn useful multilingual representations.
However, this hypothesis has been questioned by recent studies \citep{K-mBERT-ICLR-2020,conneau-etal-2020-emerging} which show that shared word-pieces only play a minor role in the performance.
These studies suggest that the model can exploit abstract structures of languages to learn shared multilingual representations.

Another line of research suggests that the learning of transferable knowledge happens even in monolingual pretraining.
\citet{artetxe-etal-2020-cross} showed that a Transformer encoder pretrained only on \Lone{} exhibits strong cross-lingual transfer performance simply by aligning the \Ltwo{} embeddings to the encoder.
\citet{papadimitriou-jurafsky-2020-learning} pretrained LSTM encoders with natural languages and non-linguistic data (\eg, code, music, and artificial data) to demonstrate that the encoders achieve reasonable performance in Spanish language modeling.
These studies provide additional evidence for the existence of transferable linguistic knowledge learned in the model.

Then what is such knowledge?
Probing studies \cite{ijcai2018-796,conneau-etal-2018-cram} have revealed that the model captures language-independent structures such as universal dependency relations \citep{chi-etal-2020-finding} and subjecthood \citep{papadimitriou-etal-2021-deep}.
However, the probing methodology does not answer whether such linguistic knowledge contributes to the performance in cross-lingual transfer.

In this study, we shed light on this question by studying transfer learning from artificial language with the Test for Inductive Bias via Language Model Transfer (TILT) \citep{papadimitriou-jurafsky-2020-learning}.
This framework enables us to assess if abstract features generalizable to \Ltwo{} (natural language) are encoded in \Lone{}.
Here we explicitly design artificial languages with some structural properties as \Lone{} to investigate their transferability.

\subsection{Studying Language Models with Artificial Language}
To study the behavior of language models, several studies have employed a specific type of {\it artificial language}: artificial variants of natural languages.
A typical experimental framework is as follows: (1) create an artificial language that differs from a natural language in one linguistic property, such as word orders \citep{sinha-etal-2021-unnatural,dufter-schutze-2020-identifying,Sinha2021MaskedLM}, scripts \citep{K-mBERT-ICLR-2020,dufter-schutze-2020-identifying,conneau-etal-2020-emerging}, or morphology \citep{ravfogel-etal-2019-studying}; (2) train or evaluate the natural/artificial language models and compare the performance to analyze the model's sensitivity to the linguistic property.

However, this methodology is limited to studying linguistic properties that are easily editable to create artificial variants and also offers limited control over the experiments.
To overcome this problem, \citet{white-cotterell-2021-examining} created artificial languages by defining their own probabilistic context-free grammars (PCFG).
As the concurrent work, \citet{Chiang2022AAAI} trained \Transformer{} encoders on artificial data with token dependencies in the sequences and showed that they perform reasonably well on the GLUE benchmark \citep{DBLP:conf/iclr/WangSMHLB19}.
In this research, we design artificial languages with certain structural properties from scratch to study knowledge transferable to natural language.

%% file: sections/02-approach.tex
\section{Approach}

\subsection{Experimental Framework}
We first describe the experimental framework used throughout this paper, the Test for Inductive Bias via Language Model Transfer (TILT) introduced by \citet{papadimitriou-jurafsky-2020-learning}.
TILT consists of pretraining and transfer steps:

\begin{enumerate}
   \item Pretrain an encoder with a pretraining task in the source language (\Lone{}). We explore pretraining with causal language modeling in \cref{sec:tilt} and masked language modeling in \cref{sec:tilt-mlm}.
   \item Transfer the encoder to the target language (\Ltwo{}) in a downstream task. As we are interested in structural prior knowledge learned in the encoder, we discard the learned \Lone{} word embeddings and initialize the embedding layer with the \Ltwo{} vocabulary. We then train the model with the encoder parameters frozen and evaluate the task performance.
\end{enumerate}

TILT reveals how transferrable the computation induced to solve the \Lone{} pretraining task is to processing \Ltwo{}.
In this study, we are interested in the transferability of certain types of structures to natural language, and thus we primarily use hand-designed artificial languages with the structural properties as \Lone{} and natural language as \Ltwo{}.

\subsection{Designing Artificial Languages}

\label{sec:artificial_language}
Artificial languages are designed to mimic a certain property of natural language.
After providing a formal definition of artificial language, we introduce several languages used in this paper.

\subsubsection{Formulation of Artificial Language}

A artificial language refers to a set of a vocabulary and algorithms to generate sequential data for pretraining.
Each language has a sentence-length distribution $p_{len}(l)$, token vocabulary $\{ w | w \in \mathcal{V} \}$, and sentence-sampling function $f(l): l \mapsto \mathcal{V}^{l}$.
The training data is generated sentence by sentence as follows: we first sample a sentence length ($l \sim p_{len}(l)$) and then sample a sequence of tokens of that length ($[w_{1}, ..., w_{l}] \sim f(l)$).

In this study, the token vocabulary $V$ simply consists of integers (or integers with a special symbol) and is not intended to correspond to a vocabulary of any natural language.
Also the sentence-length distribution $p_{len}(l)$ is fitted with a baseline dataset in each experiment.
The focus is how to design the sentence-sampling function $f(l)$.
This determines what kind of characteristics we want to encode in the artificial dataset.

\subsubsection{Modeling Word Distribution}
\label{subsec:word_distribution}

Words in natural language are distributed in non-trivial fashions.
We will study whether prior knowledge of token distribution facilitates learning from natural language.
We first present the simplest artificial language that serves as a baseline.

\minisectionNoDot{\Uniform{}} language samples each token in a sentence independently and uniformly.
Specifically, the probability of a token $w$ being sampled is

\begin{eqnarray}
  p(w) = \frac{1}{|\mathcal{V}|}.
\end{eqnarray}

However, this deviates from the token distribution of natural language.
Natural language is empirically known to follow the Zipf's law \citep{Zipf1949HumanBA}, \ie, the relation between the frequency of a word and its rank is given by $\text{frequency}(w) \propto \text{rank}(w)^{-\alpha}$.
The coefficient $\alpha$ is typically around 1, although the coefficient shows some variation according to the corpus domain \citep{Zanette2005DynamicsOT}.

\minisectionNoDot{\Zipf{}} language captures this property and samples each token $w$ from the following probability distribution assuming $\alpha=1$:

\begin{eqnarray}
  p(w) 	\propto \frac{1}{rank(w)}.
\end{eqnarray}

The two languages introduced so far generate tokens in a sentence independently.
However, words within a sentence of natural language are known to have statistical dependencies, \ie, specific cooccurrence patterns \citep{church-hanks-1989-word}.
Consider the sentence \sentence{The cat and dog are fighting over food.}
The words \word{the} and \word{cat} would cooccur much more often than by chance because \word{cat} (noun) is dependent on \word{the} (determinant); so would \word{dog} and \word{cat} because they are topically related.
The words in a sentence are usually coherent according to some syntactic and semantic dependencies.

\minisectionNoDot{\LogLinear{}} language is designed to capture this property. Inspired by the log-linear model in \citet{arora-etal-2016-latent}, tokens in a sentence $s$ are drawn from the following probability distribution:

\begin{eqnarray}
  p(w | s) 	\propto \exp(\vec{c}_{s} \cdot \vec{v}_{w}),
\end{eqnarray}

\noindent
where $\vec{c}_s$ is the discourse vector of the sentence and $\vec{v}_w$ is the word vector of the token $w$.
Intuitively, we can imagine that the discourse vector represents the {\it topic} of the sentence and determines the unigram distribution over the vocabulary \citep{Blei2003LatentDA}.
Sampling tokens this way, non-trivial cooccurrence patterns within sentences emerge in the language.

We speculate that pretraining with the \LogLinear{} language will endow the model with an inductive bias to aggregate the context in a sentence to predict the identity or property of tokens, which is likely to benefit natural language processing.

In the experiments, the word vectors $\vec{v}_w$ are initialized with the normal distribution, and the discourse vector $\vec{c}_s$ is also drawn from the normal distribution each time we generate a sentence.
We set the dimension of the word and discourse vector to 10 as we empirically find that this makes the entire token distribution close to the Zipfian distribution.

\subsubsection{Modeling Latent Dependency Structure}
\label{subsubsec:modeling_structure}
Sentences in natural language are known to have latent structures, which are often described in the form of trees \citep{Chomsky1957} or dependency graphs \citep{Melcuk1988}.
Now we consider how to endow the sampled tokens with such structures.

In this study, we adopt a dependency-based latent structure.
Words in sentences of natural language often have dependency relations and the existence of a certain word can be predictive of another word (\eg, the verb \word{am} always cooccurs with \word{I}).
We hypothesize that, pretrained on such data, language models may acquire inductive bias towards finding relations between tokens in the input, which is presumably important in processing natural language.

Inspired by \citet{papadimitriou-jurafsky-2020-learning}, we design algorithms that generate structured sentences given a set of tokens sampled with any of the strategies described in \cref{subsec:word_distribution}.
The general idea is that half of the tokens (heads) in the vocabulary are all paired with another half of tokens (tails).
A pair of head and tail can be represented in right and left brackets with the same integer (\eg{}, \token{<123}, \token{123>}).
The pairs always appear together in a sentence and express simple dependency relations.
After determining the sentence length $l \sim f(l)$, we first sample $\frac{l}{2}$ (rounded to an integer) pairs of head and tail and then arrange them with one of the following structures.

\minisectionNoDot{\FlatDependency{}} structure simply arranges the tokens randomly while keeping the right order of the brackets (\eg, [\token{<5}, \token{<84}, \token{5>}, \token{<123}, \token{123>}, \token{84>}]).
The dependency arcs are allowed to be crossed and thus often result in a non-projective dependency structure.

\minisectionNoDot{\NestingDependency{}} language, by contrast, does not allow any dependency arcs to be crossed, and the brackets are nested hierarchically (\eg, [\token{<5}, \token{<84}, \token{84>}, \token{5>}, \token{<123}, \token{123>}]).
The sentences are generated from the stack-based algorithm described in \Appendix{sec:nesting_structure}.

These structures are similar to the {\bf \Parenthesis{}} languages used to study the inductive bias of language models in \citet{papadimitriou-jurafsky-2020-learning}.
However, our \Dependency{} languages differ from them in how to represent the head and tail tokens.
In the \Parenthesis{} language, the head and tail are represented with the same token (\eg, [\token{5}, \token{84}, \token{84}, \token{5}, \token{123}, \token{123}]), which we argue deviates from the dependency structure in natural language, because in natural language, dependency relations usually hold between different words (\eg, \word{I} and \word{am}).
We will show that this difference is in fact crucial and draw a different conclusion from \citet{papadimitriou-jurafsky-2020-learning} on the importance of the nested structure (\cref{subsec:tilt-results}).

%% file: sections/03-experiment-tilt.tex
\section{Causal Language Model Pretraining with Artificial Language}
\label{sec:tilt}

In this section, we complement the study of \citet{papadimitriou-jurafsky-2020-learning}.
While they studied the inductive bias learned in LSTM encoders with some artificial languages, here we provide additional studies with the newly introduced \LogLinear{} and \Dependency{} artificial languages, and the \Transformer{} encoder.

\subsection{Experimental Setups}
\label{subsec:tile-experimental-setups}

\minisection{Task}
We study sentence-level causal (left-to-right) language modeling (CLM), where the model needs to predict the next word given the previous context in the sentence.
Note that, \citet{papadimitriou-jurafsky-2020-learning} experiment with language modeling across sentences, but we adopt sentence-level modeling because we would like to focus on the learning of sentence structures here.
As we will see in \cref{subsec:tilt-results}, we observe the same tendency in regard to the effect of artificial pretraining where we share the setups.
The task performance is measured by the average perplexity scores for each token.

\minisection{Model}
We study two encoder architectures: \LSTM{} \citep{Hochreiter1997LongSM} and \Transformer{} \citep{NIPS2017_7181}.
These architectures are known to exhibit different abilities in capturing the underlying hierarchical structure of sequential data \citep{tran-etal-2018-importance}.

The size of word embeddings is set to 300. For both \LSTM{}  and \Transformer{} encoders, the number of layers is set to 3, and the number of parameters is configured to be the same (6.9M parameters) to enable a fair comparison between architectures (for further details, see \Appendix{appendix:tilt-detail}).

\minisection{Pretraining Data}
We generate artificial corpora with three unstructured languages, which randomly arrange the tokens sampled from \Uniform{}, \Zipf{}, and \LogLinear{} languages, and four structured languages which combine the \Zipf{} sampling strategy with the structures of \FlatParenthesis{}, \NestingParenthesis{}, \FlatDependency{}, and \NestingDependency{}.

We also experiment with natural language corpora.
We create training corpora from Wikipedia dumps of \English{}, \Japanese{}, and \Spanish{}.
The sentences are tokenized with the \package{Moses} tokenizer\footnote{\url{https://github.com/moses-smt/mosesdecoder}} for English and Spanish and \package{MeCab}\footnote{\url{http://taku910.github.io/mecab/}} for Japanese.

The sentence lengths of artificial data were sampled from the empirical distribution of the \English{} Wikipedia corpus.
The size of the vocabulary $|V|$ is set to 32,000 for both artificial and natural corpora, and out-of-vocabulary words in natural language are replaced with the OOV token.
For each corpus, we sample 12.8 M sentences and train the model with one iteration over the corpus.

\minisection{Evaluation Data}
We evaluate the pretrained encoders on the Penn Treebank (PTB) corpus \citep{Marcus1993BuildingAL} with preprocessing from \citet{Mikolov2010RecurrentNN}.
Note that, when we train language models with the pretrained encoders, the parameters of the encoder are not updated and only the English word embeddings are learned from scratch (optimization details in \Appendix{appendix:tilt-optim-config}).

\subsection{Results}
\label{subsec:tilt-results}

\input{sections/figures/tilt-results}

We provide two baseline models trained on the \Ltwo{} training corpus from scratch and trained with frozen random weights in the encoder to compare with pretrained encoders.
For each configuration, we pretrain three encoders with different random seeds, and for each encoder fine-tuned three models, which results in nine models in total.
We summarize the average scores and standard deviations in \Figure{fig:tilt-results}.

{\bf The \Transformer{} encoder is more flexible than \LSTM{}.}
We start by discussing overall trends.
We observe that the \Transformer{} encoders give lower perplexity scores compared to \LSTM{} regardless of pretraining language.
This tendency is in line with the observations on the surprisingly good transferability or pretrained \Transformer{} encoders to other languages \citep{conneau-etal-2020-unsupervised}, or even other modalities \cite{Lu2021PretrainedTA,Reid2022CanWH}.
We think that this is because \Transformer{} encoders are better at aggregating and preserving the context information at each time step, as we will see in \cref{sec:probing}, presumably because the \Transformer{} architecture has self-attention and residual connections.

{\bf Natural languages are better than the artificial languages.}
As expected, pretraining with natural languages (\English{}, \Spanish{} and \Japanese{}) provides better encoders for language modeling than the artificial languages both with \LSTM{} and \Transformer{}.
However, the performance differences between natural languages seem to be negligible, indicating that there is not much difference in the way the encoders process these different languages, conforming with the observation of cross-lingual transferability of pretrained encoders \citep{artetxe-etal-2020-cross}.

{\bf The \Uniform{} and \Zipf{} languages degrade the encoders.}
Looking at the difference among unstructured languages (\Figure{fig:tilt-results:a}), \Uniform{} and \Zipf{} languages give higher perplexities than the \RandomWeights{} baseline particularly with \LSTM{}.
In hindsight, it is natural that encoders would be degraded even from random weights when trained with sequences where tokens are drawn independently from each other because the encoders are not incentivized to use contextual information and will even learn to discard the input information.
We will demonstrate this with a follow-up probing experiment in \cref{sec:probing}.

{\bf The \LogLinear{} language provides a useful inductive bias to language modeling.}
On the contrary, the \LogLinear{} language gives reasonably lower perplexities compared to \RandomWeights{} (\Figure{fig:tilt-results:a}).
This indicates that knowing the existence of statistical dependency within a sentence, or learning to predict tokens from the cooccurrence information, is a useful inductive bias even though the cooccurrence statistics is not necessarily in line with \Ltwo{}.

{\bf We do not observe the importance of the nested structure in the Parenthesis languages.}
\citet{papadimitriou-jurafsky-2020-learning} showed that \LSTM{} encoders trained on the \FlatParenthesis{} and \NestingParenthesis{} structures do not provide a significant difference in perplexity, and concluded that simple non-hierarchical head-dependent-type relations are important in \LSTM{} language processing.
A similar observation can be made in \Figure{fig:tilt-results:b}: although the \NestingParenthesis{} exhibits the lower average score, there is no significant difference between \FlatParenthesis{} and \NestingParenthesis{}  ($232.9\pm30.0$ vs. $203.8\pm7.7$, $p > 0.01$ in Welch’s t-test) with the unstable results of \FlatParenthesis{}.
Also, the trend of the average scores is reversed in \Transformer{}: the \NestingParenthesis{} exhibits the higher average score ($212.4\pm8.8$) than \FlatParenthesis{} ($191.9\pm11.8$),  which makes it difficult to draw a consistent conclusion from here.

However, {\bf the \Dependency{} languages suggest that the nested structure is actually important in language modeling.}
While the \Parenthesis{} language represents dependency relations with two identical tokens (\eg, \token{4543} and \token{4543}), our \Dependency{} language represents relations with two different tokens (\eg, \token{<4543} and \token{4543>}).
We expect that expressing dependency relations with two different tokens is closer to natural language and thus provides more viable insights into natural language.
When we compare the scores of the \Dependency{} languages, \NestingDependency{} provides the lower and more stable perplexity than \FlatDependency{} with \LSTM{} ($175.7\pm4.3$ vs. $187.2\pm10.7$) and the significantly lower score with \Transformer{} ($160.6\pm1.6$ vs. $175.7\pm4.3$, $p > 0.01$ in Welch’s t-test).
Overall, \NestingDependency{} performs best among other artificial languages, indicating our \Dependency{} language is closer to natural language and the nested structure is useful for language modeling.

%% file: sections/figures/tilt-results.tex
\begin{figure*}[t]

\centering
\subfloat[Comparison of token distributions.]{\label{fig:tilt-results:a}\includegraphics[height=4.2cm]{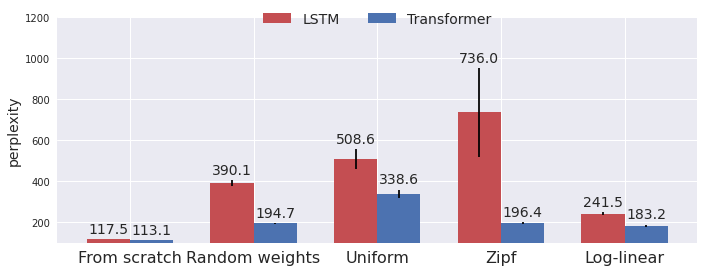}}\par\medskip
\begin{minipage}{.57\linewidth}
\centering
\subfloat[Comparison of dependency structures.]{\label{fig:tilt-results:b}\includegraphics[height=4.2cm]{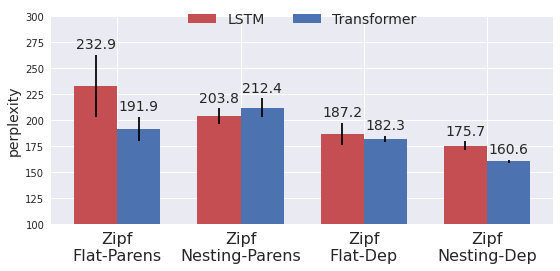}}
\end{minipage}\hfill
\begin{minipage}{.43\linewidth}
\centering
\subfloat[Comparison of natural languages.]{\label{fig:tilt-results:c}\includegraphics[height=4.2cm]{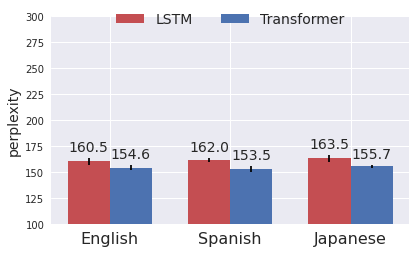}}
\end{minipage}

\caption{The perplexity scores (the lower the better) on the sentence-level causal language modeling task with the English Penn Treebank dataset. The two baselines (\FromScratch{} and \RandomWeights{}) are not pretrained, and the others are the results of pretrained encoders.}
\label{fig:tilt-results}
\vspace{-3mm}
\end{figure*}

%% file: sections/04-experiment-bert.tex
\section{Masked Language Model Pretraining with Artificial Language}
\label{sec:tilt-mlm}

We proceed to investigate transfer learning from artificial languages in one of the most successful pretraining paradigms, masked language modeling (MLM) \citep{devlin2018bert} to see if we can observe similar trends to what we see in the CLM experiment (\cref{sec:tilt}).

\subsection{Experimental Setups}
\minisection{Pretraining}
To allow for fast experimentation, we train small \Transformer{} encoders.
The size of word embeddings is set to 300 and the encoders have three layers (further details in \Appendix{appendix:mlm}).
The pretraining datasets are the same as in \cref{subsec:tile-experimental-setups}.

\minisection{Downstream Task}
We evaluate the pretrained encoders with dependency parsing to see if the structural knowledge learned with artificial language is beneficial to predict the structure of natural language.
We use the English EWT dataset from Universal Dependencies (UD) v2.8 \citep{nivre-etal-2020-universal}\footnote{\url{https://universaldependencies.org/}}.

\minisection{Model}
We adopt the biaffine graph-based parser \citep{Dozat2017DeepBA} with the \Transformer{} encoder.
The input word representations are the concatenation of word embeddings and character features computed by a character-level bi-directional LSTM encoder \citep{ling-etal-2015-finding}.
For the details on fine-tuning these models, please refer to \Appendix{appendix:mlm}.

\subsection{Results}
\label{sec:tilt-mlm-results}

\begin{figure}[t]
  \begin{center}
    \includegraphics[width=7cm]{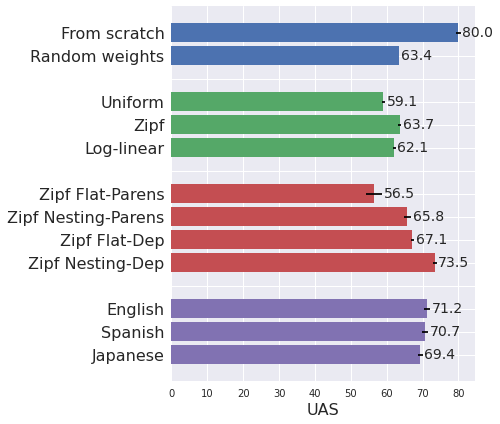}
    \caption{The downstream performance on two syntactic tasks with the English EWT dataset. The two baselines (\FromScratch{} and \RandomWeights{}) are not pretrained, and the others are the results of encoders pretrained with masked language modeling.}
    \label{fig:tilt-mlm-results}
  \end{center}
\end{figure}

We provide two baseline models trained from scratch and trained with random encoder weights.
For each pretraining language, we again train three encoders and fine-tune three models for each, and take the mean and standard deviation of the nine models.
\Figure{fig:tilt-mlm-results} shows the results.

{\bf The unstructured languages do not provide useful transferable knowledge for dependency parsing.}
The \Uniform{}, \Zipf{}, and \LogLinear{} encoders perform comparably to or worse than the \RandomWeights{} baseline.
This is in contrast with the causal language modeling task, where the \LogLinear{} language at least outperforms the \RandomWeights{} baseline (\cref{subsec:tilt-results}).

On the other hand, {\bf learning from structured languages seems to be important in \DependencyParsing{}.}
The \Dependency{} encoders outperform the \RandomWeights{} baseline, and also we can observe that learning from the nesting structure is more effective than the flat structure, and \Dependency{} languages outperform \Parenthesis{} languages, as observed in the CLM in \cref{sec:tilt}.

%% file: sections/04.5-probing.tex
\section{How much contextual information do the pretrained encoders capture?}
\label{sec:probing}

In the previous sections, we have seen that the encoders pretrained with different artificial languages exhibit various degrees of transferability to natural language.
In this section, we try to explain why pretraining with some artificial languages is better or worse for the transfer to natural language from the perspective of the amount of contextual information in the encoder outputs.

The intuition is, for example, if a pretrained encoder has learned to discard the input information, we cannot expect the encoder to perform well when transferred to any tasks.
Also, existing studies show that neural language models assign more importance to local context when they make predictions \citep{khandelwal-etal-2018-sharp,lai-etal-2020-context}.
Can we observe that encoders pretrained with artificial languages exhibit similar patterns to natural languages regarding how they encode the contextual information?

\subsection{Experimental Setups}
We investigate how much contextual information can be extracted from the outputs of the pretrained encoders by setting up a simple probing task.
In this task, the encoder is asked to recover the identity of the contextual words given the contextualized vector of a target word.

Specifically, we first randomly generate 100K sequences of integers with the length of $15\sim25$ (close to most frequent sequence lengths in the pretrained corpus) with the vocabulary size $100$ and split them into training (90K sequences), validation (5K) and test (5K) sets.

Then we simultaneously train several linear classifiers, each of which predicts the ID of the context word at a fixed relative position to the target word in the sequence, on top of a frozen pretrained encoder.
For the encoders pretrained with CLM in \cref{sec:tilt}, the target word is the last word in sequences and the classifiers predict the words at the positions of [-9, -4, -3, -2, -1, 0]; for the encoders pretrained with MLM in \cref{sec:tilt-mlm}, the target word is the middle word and the classifiers predict the words at [-6, -3, -2, -1, 0, 1, 2, 3, 6].

After training, we measure the accuracy of predicting the words at each position on the test set and interpret this as how much information on each contextual word the encoder preserves.

\input{sections/figures/probing}

\subsection{Results}
\Figure{fig:tilt-probing-results} summarizes the results of the encoders trained in \cref{sec:tilt} and \cref{sec:tilt-mlm}.

{\bf The amount of the encoded contextual information can explain the transfer performance in some obvious cases.}
In the experiment of CLM (\Figure{fig:tilt-results:a}), we observed that the \Uniform{} and \Zipf{} encoders tend to perform worse even than \RandomWeights{}.
\Figure{fig:tilt-probing-lstm:a} and \ref{fig:tilt-probing-transformer:a} demonstrate that their poor performance is because the encoders are trained to discard the input information.
The \Uniform{} and \Zipf{} encoders tend to preserve less contextual information even than \RandomWeights{} because capturing the contextual information does not lead to solving the pretraining task in these languages.

On the other hand, if words are predictable from the context, encoders are encouraged to learn to preserve the contextual information.
The \LogLinear{} encoders trained with CLM encode a decent amount of the contextual information (\Figure{fig:tilt-probing-lstm:a} and \ref{fig:tilt-probing-transformer:a}) and also performed best among the unstructured artificial languages in CLM (\Figure{fig:tilt-results:a}).
Moreover, encoders trained with natural languages (\Figure{fig:tilt-probing-lstm:c}, \ref{fig:tilt-probing-transformer:c} and \ref{fig:tilt-probing-mlm:c}) capture not only the local context well (at distance $0\sim2$) but also a modest amount of the farther context (at distance $3\sim$), which is consistent with the existing observation that LSTM encoders trained with natural language are better at memorizing the inputs than ones trained with randomly sampled data \cite{liu-etal-2018-lstms}.
In these cases, the downstream performance and the amount of the encoded contextual information seem to be correlated.

However, this trend is not as clear when comparing the structured artificial languages.
For example, the \NestingDependency{} encoders perform the best for the downstream tasks among the structured artificial languages but do not necessarily in the probing task (\Figure{fig:tilt-probing-lstm:b} and \ref{fig:tilt-probing-transformer:b}).

{\bf The nesting structure seems to facilitate encoders to remember the local context with MLM.}
The difference between the Nesting and Flat languages is striking in \Figure{fig:tilt-probing-transformer:c}.
The Nesting encoders are consistently better at capturing the local contextual information (at positions $-2\sim2$) than their flat counterparts, which may explain the better performance of the Nesting encoders in dependency parsing (\Figure{fig:tilt-mlm-results}), given that the local contextual information is particularly important to predict the syntactic characteristics of words \cite{levy-goldberg-2014-dependency,ri-tsuruoka-2020-revisiting}.

%% file: sections/figures/probing.tex
\begin{figure*}[t]

\begin{minipage}{.33\linewidth}
\centering
\subfloat[LSTM-CLM.]{\label{fig:tilt-probing-lstm:a}\includegraphics[width=5.2cm]{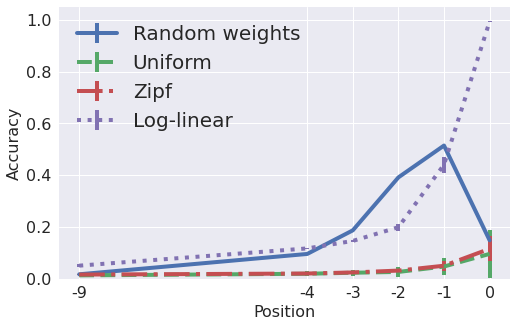}}
\end{minipage}\hfill
\begin{minipage}{.33\linewidth}
\centering
\subfloat[LSTM-CLM.]{\label{fig:tilt-probing-lstm:b}\includegraphics[width=5.2cm]{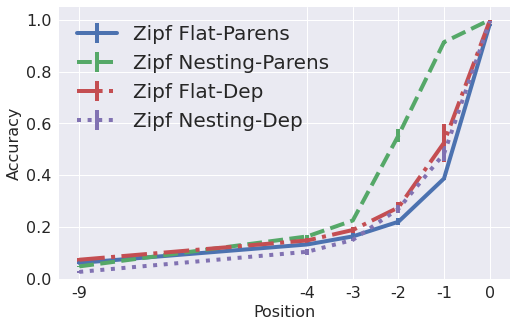}}
\end{minipage}\hfill
\begin{minipage}{.33\linewidth}
\centering
\subfloat[LSTM-CLM.]{\label{fig:tilt-probing-lstm:c}\includegraphics[width=5.2cm]{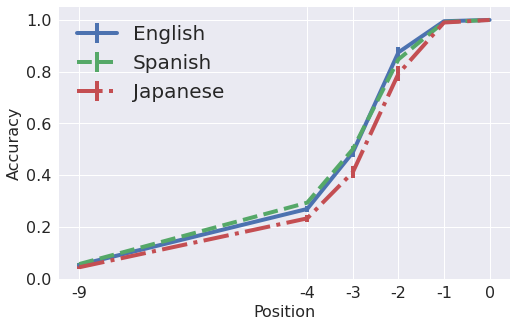}}
\end{minipage}\par\medskip 
\begin{minipage}{.33\linewidth}
\centering
\subfloat[Transformers-CLM.]{\label{fig:tilt-probing-transformer:a}\includegraphics[width=5.2cm]{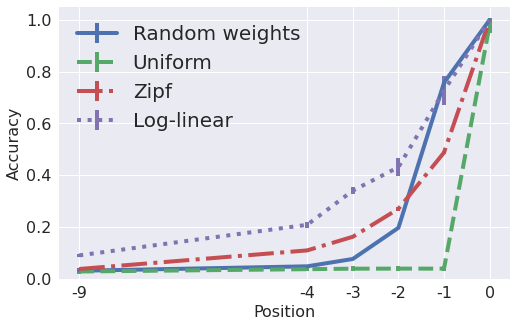}}
\end{minipage}\hfill
\begin{minipage}{.33\linewidth}
\centering
\subfloat[Transformers-CLM.]{\label{fig:tilt-probing-transformer:b}\includegraphics[width=5.2cm]{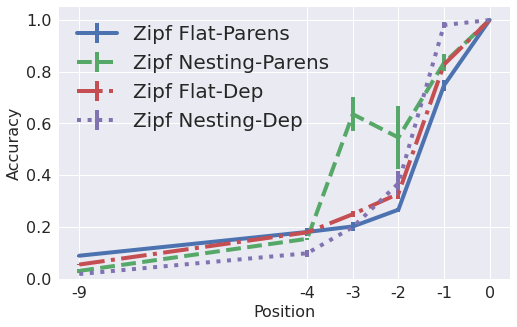}}
\end{minipage}\hfill
\begin{minipage}{.33\linewidth}
\centering
\subfloat[Transformers-CLM.]{\label{fig:tilt-probing-transformer:c}\includegraphics[width=5.2cm]{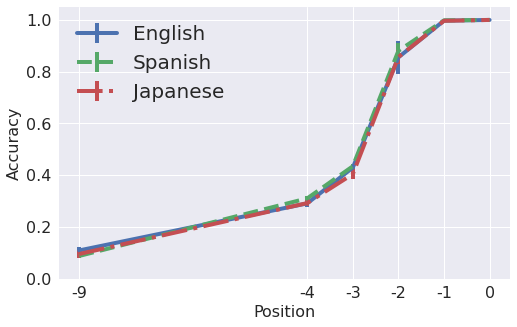}}
\end{minipage}\par\medskip 
\begin{minipage}{.33\linewidth}
\centering
\subfloat[Transformers-MLM.]{\label{fig:tilt-probing-mlm:a}\includegraphics[width=5.2cm]{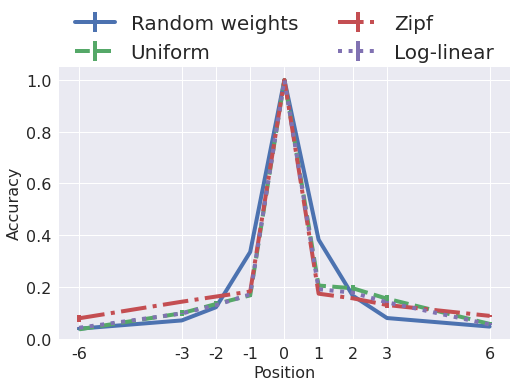}}
\end{minipage}\hfill
\begin{minipage}{.33\linewidth}
\centering
\subfloat[Transformers-MLM.]{\label{fig:tilt-probing-mlm:b}\includegraphics[width=5.2cm]{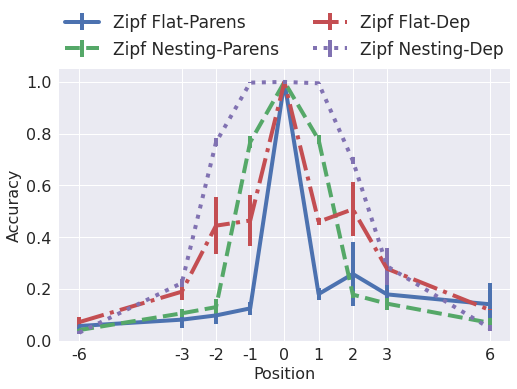}}
\end{minipage}\hfill
\begin{minipage}{.33\linewidth}
\centering
\subfloat[Transformers-MLM.]{\label{fig:tilt-probing-mlm:c}\includegraphics[width=5.2cm]{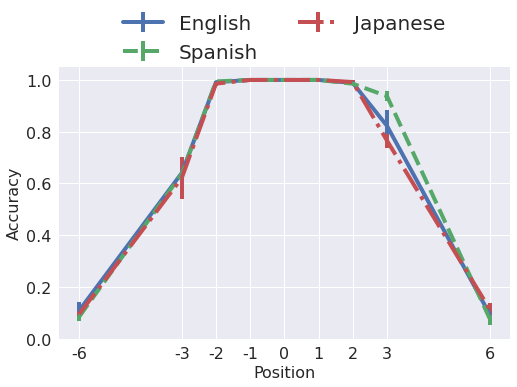}}
\end{minipage}\par\medskip

\caption{The accuracy of the task of recovering the contextual words from the encoder output of target words.}
\label{fig:tilt-probing-results}
\end{figure*}

%% file: sections/06-discussion.tex
\section{Discussion and Future Work}
In this paper, we studied what kind of structural properties in pretraining data is useful to train encoders for natural language tasks.
We have found that to achieve decent results, \Lone{} needs at least statistical dependency in a sentence (\cref{sec:tilt}), and having the head-to-tail dependency with the nesting structure is further beneficial (\cref{sec:tilt} and \cref{sec:tilt-mlm}).
The probing experiment in \cref{sec:probing} suggests that the encoders trained with languages with the above characteristics are good at capturing the positions and identities of the context words.

From these observations, we suggest a tentative answer to the initial research question: what knowledge in pretrained encoders are transferred across different languages?
That is \emph{position-aware context dependence} of language, in other words, {\bf ``tokens in a sequence can be characterized by its neighbor tokens at specific positions''}.

We think that it can explain the success of transferring the encoder across languages to some extent.
To solve natural language tasks, it is often useful to characterize words in a sentence by the words around them.
For example, to understand the semantics of a sentence, it would be useful to look for the subject by looking for a noun that precedes the word \word{is}; to parse a sentence, a word can be identified as a noun because it follows the article \word{the}.
If the encoder computes the output representation of a word in a sentence by aggregating the information from its surrounding words, that should be a useful inductive bias to solve most NLP tasks in any language.
Also, it is easy to imagine that the knowledge of position-aware context dependence gives a reasonable prior for solving sequence modeling problems in other domains, which may explain the success of cross-modality transfer of language models \citep{Lu2021PretrainedTA,Reid2022CanWH}.

Of course, we do not expect that the knowledge of position-aware context dependence explains every aspect of the success of cross-lingual transfer.
As future work, we need further investigation for a more fine-grained view of the transferred knowledge.
Important questions include how much the model size affects the transferability of the encoder or if there is any difference in the knowledge transferred among different downstream tasks.

%% file: sections/99-appendix.tex
\appendix
\section*{Appendix for ``\papertitle{}''}

\input{sections/appendices/nesting_structure}

\input{sections/appendices/tilt-details}

\input{sections/appendices/mlm-details}

\section{Computing Infrastructure}
We ran the experiments on a server with a Intel(R) Xeon(R) CPU E5-2698 v4 @ 2.20GHz CPU and 10 NVIDIA TITAN Xp GPUs.
Each pretraining and finetuning were run with a single GPU.

%% file: sections/appendices/nesting_structure.tex
\section{Generating the Nesting Structure}
\label{sec:nesting_structure}

In the Nesting languages introduced in \cref{subsubsec:modeling_structure}, tokens are ordered in a way that any dependency arcs in a sequence are not crossed.
This is realized by the stack-based algorithm in \Algorithm{algo:nesting_dependency}.
We set the probability of closing a dependency pair to 0.4 following \citet{papadimitriou-jurafsky-2020-learning}.

\newcommand{\varInputStack}{input\_pairs}
\newcommand{\varClosingStack}{closing\_stack}
\newcommand{\varSentence}{sentence}
\newcommand{\isEmpty}{is\_empty()}
\newcommand{\pop}{pop()}
\newcommand{\push}[1]{push(#1)}
\newcommand{\append}[1]{append(#1)}

\begin{algorithm}[h]
    \caption{Generating a sentence from the \NestingDependency{} language.}
    \label{algo:nesting_dependency}
    \begin{algorithmic}[1]
    \Require \varInputStack{}: Stack[($w$, $w$)]]
    \Ensure \varSentence{}: List[$w$]
    \State \varClosingStack{} = []
    \While{not \varInputStack{}.\isEmpty{}}
      \State Uniform sampling $p \sim [0, 1]$
      \If{\varClosingStack{}.\isEmpty{} or $p$ < 0.4}
        \State head, tail = \varInputStack{}.\pop{}
        \State \varSentence{}.\append{head}
        \State \varClosingStack{}.\push{tail}
      \Else
        \State tail = \varClosingStack{}.\pop{}
        \State \varSentence{}.\append{tail}
      \EndIf
    \EndWhile

    \While{not \varClosingStack{}.\isEmpty{}}
      \State tail = \varClosingStack{}.\pop{}
      \State \varSentence{}.\append{tail}
    \EndWhile

    \State \Return \varSentence{}

    \end{algorithmic}
\end{algorithm}

%% file: sections/appendices/tilt-details.tex
\section{Details of Causal Language Modeling Task}
\label{appendix:tilt-detail}

\subsection{Model configuration}
\label{appendix:tilt-model-config}

\newcommand{\modelSize}{300}
\newcommand{\lstmHiddenSize}{294}
\newcommand{\numLayers}{3}

For the experiment with causal language modeling (\cref{sec:tilt}), we set the number of layers of the LSTM and Transformer encoders to \numLayers{} and configure them so that they have the same number of parameters (2.1 M parameters without the embedding and output projection layers).
The details of configuration are shown in \Table{tb:lstm-config} and \Table{tb:transformer-config}.

The weights of the output projection layer are tied with the word embedding layer \citep{press-wolf-2017-using}.
Note that, to enable this, the LSTM encoder has an additional linear layer to project the hidden vector (\lstmHiddenSize{} dim) to the input size (\modelSize{} dim), which the Transformer encoder does not have.

\begin{table}[ht]
    \centering

    \begin{tabular}{lc} \toprule
    \# of layers & \numLayers{}   \\
    input size   & \modelSize{} \\
    hidden size  & \lstmHiddenSize{} \\
    \bottomrule
    \end{tabular}

    \caption{Configuration of the LSTM encoder.}
    \label{tb:lstm-config}
\end{table}

\begin{table}[ht]
    \centering

    \begin{tabular}{lc} \toprule
    \# of layers          & \numLayers{}   \\
    size                  & \modelSize{}  \\
    feedforward size      & 600 \\
    \# of attention heads & 4 \\
    \bottomrule
    \end{tabular}

    \caption{Configuration of the Transformer encoder.}
    \label{tb:transformer-config}
\end{table}

\subsection{Optimization}
\label{appendix:tilt-optim-config}

We optimize the pretrained models for 10k steps with 12.8 M sentences and the batch size of 128 using AdamW \citep{DBLP:conf/iclr/LoshchilovH19}.
We use the the Noam Learning rate scheduler described in \citet{NIPS2017_7181} with the warmup steps of 4000, and the other hyper-parameter details are shown in Table \ref{tb:pretraining-config}.
We use the same hyper-parameters for fine-tuning with the \Ltwo{} language.

\begin{table}[ht]
    \centering
    \begin{tabular}{lc}
        \toprule
        Name                                  & Value \\ \midrule
        Pretraining minimum sentence length               & 6 \\
        Pretraining maximum sentence length               & 60 \\
        Dropout                               & 0.1   \\
        Weight decay                          & 0.01   \\
        Adam $\beta_1$                        & 0.9    \\
        Adam $\beta_2$                        & 0.98  \\
        Adam $\epsilon$                       & 1e-9   \\
        Gradient clipping                     & 0.25 \\
        \bottomrule
    \end{tabular}
    \caption{Hyper-parameters for pretraining.}
    \label{tb:pretraining-config}
\end{table}

%% file: sections/appendices/mlm-details.tex
\section{Details of Masked Language Modeling Task}
\label{appendix:mlm}

\subsection{Model configuration}
\label{appendix:tilt-mlm-model-config}

\newcommand{\modelSizeMLM}{300}
\newcommand{\numLayersMLM}{3}

For the experiment with masked language modeling (\cref{sec:tilt-mlm}), we set the number of layers of the Transformer encoders to \numLayersMLM{}. The details of configuration are shown in \Table{tb:transformer-mlm-config} (2.1 M parameters without the embedding and output projection layers).

The hyper-parameters for the masked language modeling task is shown in \Table{tb:mlm-config}.
For optimization, we used the same hyper-parameters as in \Appendix{appendix:tilt-optim-config}.

\begin{table}[ht]
  \centering
  \begin{tabular}{lc} \toprule
  \# of layers          & \numLayersMLM{}   \\
  size                  & \modelSizeMLM{}  \\
  feedforward size      & 600 \\
  \# of attention heads & 4 \\
  \bottomrule
  \end{tabular}
  \caption{Model configuration of the Transformer encoder.}
  \label{tb:transformer-mlm-config}
\end{table}

\begin{table}[ht]
    \centering
    
    \begin{tabular}{lc} \toprule
      Mask probability for words            & 15\%  \\
      Random-word probability for words     & 10\%  \\
      Unmasked probability for words        & 10\%  \\
    \bottomrule
    \end{tabular}
    \caption{The hyper-parameters for masked language modeling.}
    \label{tb:mlm-config}
\end{table}